\documentclass[twoside,11pt]{article}

\usepackage{jmlr2e}

\usepackage[greek,english]{babel}
\usepackage[utf8]{inputenc} 
\usepackage[T1]{fontenc}    
\usepackage{hyperref}       
\usepackage{url}            
\usepackage{booktabs}       
\usepackage{nicefrac}       
\usepackage{microtype}      
\usepackage[htt]{hyphenat}
\usepackage[export]{adjustbox}
\usepackage{float}
\usepackage{bbm,bm}
\usepackage{mdframed}
\usepackage{relsize}
\usepackage{tikz}
\usepackage{pgfplots}
\usepackage{caption,subcaption}
\usepackage{multirow}
\usepackage{xcolor}
\usepackage{dsfont}
\usepackage{amsmath,amssymb,amsfonts}

\usepackage{listings}

\definecolor{codegreen}{rgb}{0,0.6,0}
\definecolor{codegray}{rgb}{0.5,0.5,0.5}
\definecolor{codepurple}{rgb}{0.58,0,0.82}
\definecolor{backcolour}{rgb}{0.95,0.95,0.92}

\lstdefinestyle{mystyle}{
    backgroundcolor=\color{backcolour},   
    commentstyle=\color{codegreen},
    keywordstyle=\color{magenta},
    numberstyle=\tiny\color{codegray},
    stringstyle=\color{codepurple},
    basicstyle=\ttfamily\footnotesize,
    breakatwhitespace=false,         
    breaklines=true,                 
    captionpos=b,                    
    keepspaces=true,                 
    numbers=left,                    
    numbersep=5pt,                  
    showspaces=false,                
    showstringspaces=false,
    showtabs=false,                  
    tabsize=2
}

\usepackage{algorithm}
\usepackage{algorithmic}

\newcommand{\bb}{\texttt{BackboneLearn}}

\ShortHeadings{BackboneLearn}{Digalakis Jr and Ziakas}
\firstpageno{1}

\begin{document}

\title{BackboneLearn: A Library for Scaling Mixed-Integer Optimization-Based Machine Learning}

\author{\name Vassilis Digalakis Jr \email digalakis@hec.fr
       \AND
       \name Christos Ziakas \email chziakas@gmail.com \\
       \addr Department of Information Systems and Operations Management, HEC Paris}

\editor{}

\maketitle

\begin{abstract}
We present \bb: an open-source software package and framework for scaling mixed-integer optimization (MIO) problems with indicator variables to high-dimensional problems. This optimization paradigm can naturally be used to formulate fundamental problems in interpretable supervised learning (e.g., sparse regression and decision trees), in unsupervised learning (e.g., clustering), and beyond;
\bb\ solves the aforementioned problems faster than exact methods and with higher accuracy than commonly used heuristics. The package is built in Python and is user-friendly and easily extensible: users can directly implement a backbone algorithm for their MIO problem at hand. The source code of \bb\ is available on GitHub\footnote{Link: \url{https://github.com/chziakas/backbone_learn}}.
\end{abstract}


\section{Introduction}
Mixed-integer optimization (MIO) problems with indicator variables are encountered in a wide array of applications ranging from (sparse) machine learning (ML) to network design and portfolio selection. For example, given data $\boldsymbol{X} \in \mathbb{R}^{n\times p}$, responses $\boldsymbol{y} \in \mathbb{R}^{n}$, and a desired sparsity level $k\ll p$, the sparse linear regression is formulated as $\min_{\boldsymbol{\beta}} \|\boldsymbol{y} - \boldsymbol{X} \boldsymbol{\beta} \|_2^2 \ \text{s.t.} \|\boldsymbol{\beta}\|_0 \leq k$; then, each regression coefficient $\beta_j$ (and the corresponding feature $j$) is paired with an indicator $z_j = \mathbbm{1}_{\beta_j \neq 0}$. Over the past decade, MIO-based methods for ML \citep{bertsimas2019machine,gambella2021optimization} have scaled from being intractable beyond toy scales (e.g., sparse regression/decision tree problems with a few hundred/a dozen features) to now being able to handle much more realistically-sized problems: see, e.g., \cite{hazimeh2022sparse,bertsimas2021slowly} for sparse regression and \cite{aghaei2021strong,mazumder2022quant} for decision trees. 
The backbone framework, introduced by \cite{bertsimas2022backbone}, is a general, heuristic framework (albeit with some theoretical guarantees) to scale such techniques even further: we solved, in minutes, sparse regression problems with tens of millions of features and decision tree problems with hundreds of thousands. 

Besides the original work by \cite{bertsimas2022backbone}, the backbone framework has been successfully implemented in vehicle routing \citep{bertsimas2019online} and compressed sensing \citep{bertsimas2023compressed}. Backbone-type algorithms have been shown to be relevant in supervised ML, e.g., sparse SVM \citep{bi2003dimensionality}, in unsupervised ML, e.g., sparse PCA \citep{cory2022sparse}, in magnet optimization \citep{kaptanoglu2022permanent}, in learning sparse nonlinear dynamics \citep{bertsimas2023learning}, and, more recently, for mechanistic interpretability of large language models \citep{gurnee2023finding}. Lastly, exact, safe screening-based approaches, such as the works of \cite{atamturk2020safe,deza2022safe}, can also be described and implemented under the backbone framework.

In this paper, we describe \bb, an open-source software package that seeks to help data science practitioners \textbf{(i)} use the algorithms developed by \cite{bertsimas2022backbone} and extensions thereof (e.g., we develop a novel backbone algorithm for clustering) in combination with state-of-the-art solvers; \textbf{(ii)} easily implement their own \texttt{BackboneLearner} for their problem at hand (in supervised ML, unsupervised ML, and beyond); \textbf{(iii)} interface with various state-of-the-art MIO-based interpretable ML algorithms and packages, including the \texttt{L0Learn} \citep{hazimeh2023l0learn} and \texttt{L0Bnb} \citep{hazimeh2022sparse} packages for sparse regression and the \texttt{ODTLearn }\citep{vossler2023odtlearn} package for optimal decision trees (in other words, \bb\ can be used as a joint feature selector and model fitter for the aforementioned methodologies). Overall, this project aims to bring together under a single framework and software package all the backbone-type approaches described in the previous paragraph, to facilitate adoption by practitioners; we believe this is especially relevant following the increased interest in and the established benefits of developing exact solutions for interpretable ML models. 

\section{The Backbone Framework}

The backbone framework, upon which \bb\ is built, operates in two phases: we first extract a “backbone set” of potentially ``relevant indicators'' (i.e., indicators that are nonzero in the optimal solution) by solving a number of specially chosen, tractable subproblems; we then use traditional techniques to solve a reduced problem to optimality or near-optimality, considering only the backbone indicators. A screening step often precedes the first phase, to discard indicators that are almost surely irrelevant. \cite{bertsimas2022backbone} develop backbone methods for various supervised ML models (whereby each indicator corresponds to a feature), including sparse linear and logistic regression, and classification trees, all of which are implemented within \bb; in fact, for the case of sparse linear regression, their theoretical analysis shows that, under certain conditions and with high probability, the backbone set consists only of the truly relevant indicators. Extending their original work, \bb\ also implements a backbone algorithm for clustering and sets the foundation for a series of additional extensions. 

\paragraph{The Backbone Algorithm. } In Algorithm \ref{alg:backbone}, we provide a general, modularized backbone algorithm. The input consists of \textbf{(i)} the data $D$ (e.g., in supervised learning, $D=(\boldsymbol{X},\boldsymbol{y})$, whereas in unsupervised learning $D=\boldsymbol{X}$), \textbf{(ii)} a set of hyperparameters including the number of subproblems, the subproblem size, the number of features to be discarded through a screening test or ``proxy,'' and the maximum allowable backbone size, and \textbf{(iii)} a set of application-specific functions. The latter includes the \texttt{screen} function that discards irrelevant indicators after calculating their utilities (denoted by $s$), the \texttt{construct\_subproblems} function that builds a series of tractable subproblems (decreasing in number in each backbone iteration $t$), and the \texttt{fit\_subproblem} and \texttt{fit} functions that, respectively, solve the subproblems (possibly heuristically) and the ``reduced'' problem (considering only the indicators that have been selected in the backbone set $\mathcal{B}$). The output is the learned model (e.g., sparse regression coefficients; decision tree parameters; cluster assignment).

\begin{algorithm*}
\caption{General Backbone Algorithm}
\label{alg:backbone}
\begin{algorithmic}
    \REQUIRE Data $D$, hyperparameters $(M,\beta,\alpha,B_{\max})$, functions (\verb|screen|, \verb|construct_subproblems|, \verb|fit_subproblem|, \verb|extract_relevant|, \verb|fit|).
    \ENSURE Learned model.
    \STATE $t \leftarrow 0$; \ $\mathcal{U}_0, \*s \leftarrow \verb|screen|(D,\alpha)$
    \REPEAT
        \STATE $\mathcal{B} \leftarrow \emptyset$
        \STATE $(\mathcal{P}_m)_{m\in[M]} \leftarrow \verb|construct_subproblems|(\mathcal{U}_t,\*s,\lceil \frac{M}{2^t} \rceil,\beta)$
        \FOR{$m \in [M]$}
            \STATE $\text{model}_m \leftarrow \verb|fit_subproblem|(D, {\mathcal{P}_m})$
            \STATE $\mathcal{B} \leftarrow \mathcal{B} \cup \verb|extract_relevant|(\text{model}_m)$
        \ENDFOR
        \STATE $t \leftarrow t+1$; \ $\mathcal{U}_t \leftarrow \mathcal{B}$
    \UNTIL{$|\mathcal{B}| \leq B_{\max}$ (or other termination criterion is met)}
    \STATE model $\leftarrow$ \verb|fit|$(D, {\mathcal{B}})$
\end{algorithmic}
\end{algorithm*}

\paragraph{Backbone Examples. } We proceed by giving more concrete examples of backbone sets for specific problems. For sparse linear regression, the backbone set can be written as:
\begin{equation*}
    \label{eqn:sr_backbone_construction}
    \begin{split}
        \mathcal{B} = \bigcup_{m=1}^M 
        \left\{ j: \beta_j^{(m)} \not= 0, 
        \quad \boldsymbol{\beta}^{(m)} = \text{argmin}_{\substack{ \boldsymbol{\beta} \in \mathbb{R}^{p}:\\ \Vert \boldsymbol{\beta} \Vert_0 \leqslant k_m,\\ {\beta}_j=0 \ \forall j \not\in \mathcal{P}_m}}
        \quad \| \boldsymbol{y} - \boldsymbol{X}\boldsymbol{\beta} \|_2^2 \right\}.
    \end{split}
\end{equation*}
The solution to each subproblem $\boldsymbol{\beta}^{(m)}$ can be solved using fast, heuristic methods. After $\mathcal{B}$ is computed, we solve (to optimality) the original problem under the additional constraint ${\beta}_j = 0, \quad \forall j \not\in \mathcal{B}$. For optimal decision trees, $\mathcal{B}$ consists of all features that have not been selected in any split node in any subproblem or have small importance across all subproblems.

For clustering, the goal is to group the $n$ data points in $\boldsymbol{X}$ into $k$ clusters $\{S_1, ..., S_k\}$ to minimize $\sum_t \frac{1}{|S_t|} \sum_{\substack{i<j: \\ \boldsymbol x_i,\boldsymbol x_j \in S_t}} \| \boldsymbol x_i-\boldsymbol x_j\|_2^2$. To develop a clustering backbone algorithm, we rely on the formulation by \cite{grotschel1989cutting}, who view the clustering problem as a graph clique partitioning problem and represent each cluster as a fully connected component. Then, \bb\ aims to sparsify the graph that contains all feasible clique partitionings.
Formally, let $\mathcal{Z} = \left\{ \boldsymbol z, \boldsymbol{\zeta}: \quad \sum_t z_{it} = 1 \ \forall i, \quad \sum_i z_{it} \geq b \ \forall t, \quad \zeta_{ijt} = z_{it}z_{jt} \ \forall i,j,t \right\}$, where $z_{it}$ (respectively $\zeta_{ijt}$) indicates whether point $i$ is put into cluster $t$ (points $i,j$ are put together into cluster $t$), and $f(\boldsymbol \zeta; \boldsymbol X) = \sum_{i,j,k} \zeta_{ijk} \| \boldsymbol x_i-\boldsymbol x_j\|^2$.
The backbone set can be expressed as:
\begin{equation*}
\label{eqn:clustering_backbone_construction}
\begin{split}
    \mathcal{B} = \bigcup_{m=1}^M 
    \left\{ (i,j): \sum_k \zeta_{ijk}^{(m)} = 1, 
    \quad \*\zeta^{(m)} = \text{argmin}_{(\*z,\*\zeta \in \mathcal{Z})} f(\*\zeta; \*X^{(m)}) \right\}.
    \end{split}
\end{equation*}
After $\mathcal{B}$ is computed, we solve (to optimality) the original problem under the additional constraint $z_{it}+z_{jt} \leq 1 \ \ \forall (i,j) \not\in \mathcal{B} \ \forall t$. To further scale the algorithm, we linearize the bilinear constraint and include additional ones (e.g., we also encode the set $\mathcal{B}^\complement$).

\section{The BackboneLearn Package}
\paragraph{Implementation, Usage, and Extensibility. } \bb\ is implemented in Python 3.9 and can be installed by executing \texttt{pip install backbone-learn}. We use \textbf{(i)} Poetry for reproducibility and to automatically manage dependencies, \textbf{(ii)} continuous integration with the Pytest testing framework, and \textbf{(iii)} GitHub Actions; \bb\ achieved the highest rating for code quality by the code analysis tool Codacy. We interface \bb\ with state-of-the-art, open-source packages: for sparse regression, we rely on \texttt{L0Learn} \citep{hazimeh2023l0learn}, \texttt{L0Bnb} \citep{hazimeh2022sparse}, and \texttt{GLMNet} \citep{zou2005regularization}; for decision trees, we utilize \texttt{ODTLearn }\citep{vossler2023odtlearn} and \texttt{ScikitLearn} \citep{pedregosa2011scikit}; for mixed-integer optimization, we use the open-source solver \texttt{Cbc} \citep{forrest2005cbc} and the Python optimization modeling library \texttt{PuLP} \citep{mitchell2011pulp}.

The \bb\ documentation provides extensive examples for sparse regression, decision trees, clustering, and extensions. \bb\ for sparse regression is run using:
\begin{lstlisting}[language=Python,numbers=none]
bb = BackboneSparseRegression(alpha=0.5, beta=0.5, num_subproblems=5, lambda_2=0.001, max_nonzeros=10) # Initialize BackboneSparseRegression
bb.fit(X, y) # Fit the model
y_pred = bb.predict(X) # Make predictions
\end{lstlisting}

To implement custom backbone algorithms, \bb\ offers two primary classes for extension: \texttt{BackboneSupervised} for supervised ML and \texttt{BackboneUnsupervised} for unsupervised ML. Thus, a \texttt{CustomBackboneAlgorithm} requires implementing the \texttt{set\_solvers()} method within the selected subclass, which, in turn, defines the algorithm's core components: \textbf{(i)} the \texttt{CustomScreenSelector} (optional screening method) including the \texttt{calculate\_utilities()} function, \textbf{(ii)} the \texttt{CustomHeuristicSolver} (subproblem solution method) with the \texttt{get\_relevant()} and \texttt{fit()} functions, and \textbf{(iii)} the \texttt{CustomExactSolver} (reduced problem solution method) with the \texttt{fit()} and \texttt{predict()} functions. 

\paragraph{Experiments. } Our numerical experiments investigate the following questions: does \bb\ speed up exact methods, does it enhance the accuracy of fast heuristics, and is it consistent across different problems? All experiments were performed on a standard Apple M2 @ 3.50GHz running macOS Ventura release 13.4.1.(c) with 8Gb of RAM. The code for replicating our simulations is available on GitHub. Here, we limit our analysis to synthetic experiments; the practical relevance of the backbone framework has been documented in full-scale, real-world experiments and problems in \cite{bertsimas2022backbone,bertsimas2019online,bertsimas2023compressed}. 

In \textit{sparse regression}, we generate synthetic data as per a ground truth sparse linear regression model under the fixed design setting (following \cite{hazimeh2022sparse}); we compute the full regularization path for \texttt{GLMNet} and \texttt{L0Bnb} both when used independently and as part of \bb; we evaluate each method's (\texttt{GLMNet}, \texttt{L0Bnb}, \bb) accuracy (using the R$^2$ statistic) and total computational time. In \textit{decision trees}, we generate binary classification data by evenly distributing a set of normally distributed clusters among classes and adding noise and feature interdependence; we cross-validate hyperparameters (e.g., tree depth); we evaluate each method's (\texttt{CART}, \texttt{ODTLearn}, \bb) accuracy (using the AUC) and total computational time. In \textit{clustering}, we generate noisy isotropic Gaussian blobs and, to create ambiguity, we assume the target number of clusters is greater than the true number; we evaluate each method's (\texttt{KMeans}, \texttt{Exact}, \bb) accuracy (using the Silhouette score) and total computational time. Below each method, we specify the problem size, defined in terms of the number of data points ($n$), features ($p$), and true relevant features or target clusters ($k$); we pick the problem sizes so that the exact methods are at their limits within the allowed one-hour budget that we impose.

Our results, detailed in Table \ref{tab:results} and averaged over 10 repetitions, verify the improved performance of \bb\ in terms of \textbf{(i)} significantly reducing the solution times of exact methods and \textbf{(ii)} improving upon the performance of heuristics. \bb\ consistently recovers the optimal solution (which, given our data generating process, is known) and achieves provable optimality (with suboptimality gaps under 1\%) --- albeit in reduced problems. Regarding runtime, in sparse regression \bb\ performs best for larger values of $\alpha,\beta$, suggesting that, when possible, it is preferred to solve larger subproblems that include more signal; for decision trees, efficiency increases with smaller subproblems, in agreement with the known benefits of feature sampling in random forest feature sampling; lastly, for clustering, the effect of \bb's hyperparameters is negligible (the method effectively selects the best clustering among the ones examined in subproblems).
\begin{table}[ht!]
    \caption{Numerical Results.}
    \label{tab:results}
    \centering
    \resizebox{0.75\textwidth}{!}{%
        \begin{tabular}{llrrrrrr}
            \toprule
            \textbf{Problem} & \textbf{Method} & \textbf{M} & \textbf{a} & \textbf{b} & \textbf{Accuracy} & \textbf{Time (sec)} & \textbf{Backbone Size} \\
            \midrule
            \multirow{6}{*}{\shortstack[l]{Sparse\\Regression\\$(n,p,k)=$\\$(500,5000,10)$}} & \texttt{GLMNet} & - & - & - & 0.871 & 15 & - \\
            & \texttt{L0Bnb} & - & - & - & 0.883 & 1175 & - \\
            & \multirow{4}{*}{\texttt{BbLearn}} & 5 & 0.1 & 0.5 & 0.884 & 483 & 376 \\
            & & 5 & 0.5 & 0.9 & 0.884 & 94 & 40 \\
            & & 10 & 0.1 & 0.5 & 0.884 & 692 & 449 \\
            & & 10 & 0.5 & 0.9 & 0.884 & 202 & 87 \\
            \midrule
            \multirow{6}{*}{\shortstack[l]{Decision\\Trees\\$(n,p,k)=$\\$(500,100,10)$}} & \texttt{CART} & - & - & - & 0.708 & 1 & - \\
            & \texttt{ODTLearn} & - & - & - & 0.639 & 3600 & - \\
            & \multirow{4}{*}{\texttt{BbLearn}} & 5 & 0.1 & 0.5 & 0.714 & 34 & 9 \\
            & & 5 & 0.5 & 0.9 & 0.714 & 2285 & 42 \\
            & & 10 & 0.1 & 0.5 & 0.714 & 108 & 10 \\
            & & 10 & 0.5 & 0.9 & 0.714 & 3137 & 46 \\
            \midrule
            \multirow{4}{*}{\shortstack[l]{Clustering\\$(n,p,k)=$\\$(200,2,5)$}} & \texttt{KMeans} & - & - & - & 0.454 & 0 & - \\
            & \texttt{Exact} & - & - & - & - & 3600 & - \\
            & \multirow{2}{*}{\texttt{BbLearn}} & 5 & - & 1.0 & 0.481 & 116 & 3483 \\
            & & 10 & - & 1.0 & 0.481 & 111 & 3483 \\
            \bottomrule
        \end{tabular}%
    }
\end{table}

\vspace{-20pt}

\section{Conclusion}

We introduced \bb, an open-source software package and framework for scaling MIO problems with indicator variables to high-dimensional problems, addressing challenges in supervised and unsupervised learning, and beyond. \bb\ currently offers algorithms for sparse regression, decision trees, and clustering, and is easily extensible: users can directly implement a backbone algorithm for their MIO problem at hand. Our numerical experiments demonstrate \bb's ability to enhance the efficiency of exact methods and improve upon the accuracy of commonly used heuristics, validating the practical relevance of backbone-type algorithms. We aim for the \bb package to evolve into a hub for highly scalable MIO-based ML algorithms.



\bibliography{main}

\end{document}